\documentclass[10pt,journal,compsoc]{IEEEtran}

\usepackage{array}
\usepackage{algorithm}
\usepackage{algpseudocode}
\usepackage{amsmath}
\usepackage{amssymb}
\usepackage{amsfonts}
\usepackage{booktabs}
\usepackage{epsfig}
\usepackage{epstopdf}
\usepackage{enumerate}
\usepackage{float}
\usepackage{multicol}
\usepackage{multirow}
\usepackage{ntheorem}
\begin{document}
\newtheorem{theorem}{Theorem}
\newtheorem{assumption}{Assumption}
\newtheorem{proposition}{Proposition}
\newtheorem{definition}{Definition}
\newtheorem{lemma}{Lemma}
\newtheorem{corollary}{Corollary}
\newtheorem{remark}{Remark}
\newtheorem{construction}{Construction}
\newtheorem{problem}{Problem}
\newtheorem{alg}{Algorithm}[section]

\title{Covert Model Poisoning Against Federated Learning: Algorithm Design and Optimization}

\author{Kang~Wei,~\IEEEmembership{Student Member,~IEEE,}
        Jun~Li,~\IEEEmembership{Senior Member,~IEEE,}
        Ming~Ding,~\IEEEmembership{Senior Member,~IEEE,}
        Chuan~Ma,
        Yo-Seb Jeon,~\IEEEmembership{Member,~IEEE,}
        and~H.~Vincent~Poor,~\IEEEmembership{Life~Fellow,~IEEE}
\IEEEcompsocitemizethanks{\IEEEcompsocthanksitem Kang~Wei, Jun~Li and Chuan~Ma are with School of Electrical and Optical Engineering, Nanjing University of Science and Technology, Nanjing, China. Jun~Li is also with the School of Computer Science and Robotics, National Research Tomsk Polytechnic University, Tomsk, 634050, RUSSIA. E-mail: \{kang.wei, jun.li, chuan.ma\}@njust.edu.cn.
\IEEEcompsocthanksitem Ming~Ding is with Data61, CSIRO, Sydney, Australia. E-mail: ming.ding@data61.csiro.au.
\IEEEcompsocthanksitem Yo-Seb Jeon is with Department of Electrical Engineering, Pohang University of Science and Technology (POSTECH), Pohang, South Korea. E-mail: yoseb.jeon@postech.ac.kr.
\IEEEcompsocthanksitem H.~Vincent~Poor is with Department of Electrical Engineering, Princeton University, NJ, USA. E-mail: poor@princeton.edu.}
\thanks{Manuscript received April 19, 2005; revised August 26, 2015.}}

\IEEEtitleabstractindextext{%
\begin{abstract}
Federated learning (FL), as a type of distributed machine learning frameworks, is vulnerable to external attacks on FL models during parameters transmissions.
An attacker in FL may control a number of participant clients, and purposely craft the uploaded model parameters to manipulate system outputs, namely, model poisoning (MP).
In this paper, we aim to propose effective MP algorithms to combat state-of-the-art defensive aggregation mechanisms (e.g., Krum and Trimmed mean) implemented at the server without being noticed, i.e., covert MP (CMP).
Specifically, we first formulate the MP as an optimization problem by minimizing the Euclidean distance between the manipulated model and designated one, constrained by a defensive aggregation rule.
Then, we develop CMP algorithms against different defensive mechanisms based on the solutions of their corresponding optimization problems.
Furthermore, to reduce the optimization complexity, we propose low complexity CMP algorithms with a slight performance degradation.
In the case that the attacker does not know the defensive aggregation mechanism, we design a blind CMP algorithm, in which the manipulated model will be adjusted properly according to the aggregated model generated by the unknown defensive aggregation.
Our experimental results demonstrate that the proposed CMP algorithms are effective and substantially outperform existing attack mechanisms.
\end{abstract}

}
\maketitle

\IEEEdisplaynontitleabstractindextext

%
\IEEEpeerreviewmaketitle

\section{Introduction}
With the development of Internet of Things (IoT), various end devices, such as sensors and smart phones, generate huge amounts of data and send them to cloud servers for processing~\cite{ Cai2017IoT}.
Big data-driven artificial intelligence (AI) has been widely applied in many aspects of modern society~\cite{Mohammadi2018Deep,Nguyen2020Wireless}.
As a result, data privacy and confidentiality have become more and more concerned as they usually contain clients' sensitive information~\cite{ma2020rdpgan,Shaham2020Privacy,wei2019performance}.
Federated learning (FL), emerging as a promising distributed machine learning paradigm~\cite{wang2019adaptive}, is capable of pushing model training to end devices without exposing their private training data.
Therefore, in recent years, a wide range of privacy-sensitive applications are developed along with FL, such as mobile keyboard prediction~\cite{Andrew2018Federated} and visual object detection for safety~\cite{Yang2019Federated}, etc.

Although FL can help preserve clients' privacy, it is possibly trained across a fleet of unreliable devices with private and uninspectable datasets~\cite{ma2019safeguarding,Wang2019beyond,Yu2019differentially,Melis2019Exploiting,li2019federated}, compared with distributed datacenter learning and centralized learning schemes.
Therefore, a new attack framework on federated training systems has be explored~\cite{bagdasaryan2018backdoor}, i.e. model poisoning attack.
Model poisoning takes advantage of the observation that a participant client in FL can directly influence parameters of the joint model.
Therefore, in model poisoning, the attacker may take over a number of clients and manipulates the local model parameters sent from these clients to the server during the learning process~\cite{kairouz2019advances, fang2019local}.
For example, with model poisoning, a competitor can degrade performance of a FL model or achieve its own goals by a particular trojan trigger~\cite{fang2019local,Mu2017Towards}.
In addition, the client compromised by an attacker can also incorporate the evasion of potential defenses into its loss function during training process.

Therefore, defence mechanisms have drawn more and more attentions.
Detection methods based on model validation are proposed to capture anomalous models uploaded by the clients, and reduce the weighting of these models when performing aggregation~\cite{Zhao2020Privacy,Zhao2020Shielding,Ying2020PDGAN}.
However, these detection methods, relying on auxiliary validation dataset, will increase the risk of the privacy leakage and become impractical for real-time training due to high complexity.
As a type of alternative defence mechanisms, robust aggregation rules (e.g., Krum~\cite{Blanchard2017} and Trimmed mean~\cite{Dong2019Byzantine}) take the advantage of low complexity and no additional privacy concerns.
To be specific, in Krum~\cite{Blanchard2017}, for each client's model, the server will calculate the sum of its Euclidean distances to the models of other clients, and select the one which has the minimum sum.
In Trimmed mean~\cite{Dong2019Byzantine}, for the parameters embedded in a designated position of local models, the server will remove a number of the largest and smallest values before aggregation.
It can be noted that both Krum and Trimmed mean can effectively mitigate the impact of unreasonable models with low complexity.

In this paper, we are interested in proposing model poisoning attacks against state-of-the-art robust aggregation rules implemented at the server.
The proposed attacking algorithms will stealthily cheat the server to adopt compromised models from the manipulated clients, termed as covert model poisoning (CMP).
It is expected that the designed CMP will destroy the original FL model for performance degradation or achieve the attacker's purposes.
To the best of the authors' knowledge, this the first piece of work in FL that systematically studies on model poisoning.

The main contributions can be summarized as follows:
\begin{itemize}
\item We formulate the model poisoning as an optimization   problem by minimizing the Euclidean distance between the manipulated model and designated one, constrained by a            defensive aggregation rule.
    Then, we develop CMP algorithms against different defensive aggregation rules according to the solutions of their corresponding optimization problems.
\item We also propose a low complexity CMP algorithm for Krum     with a slight performance degradation.
    In this algorithm, we reduce the searching dimension of the optimization problem when comparing the summations of
    Euclidean distances among the clients.
\item  In the case that the attacker does not know the defence mechanism, we design a blind CMP algorithm, in which the manipulated model will be adjusted properly according to the aggregated model.
\item We conduct extensive experiments on real-word datasets, i.e., MNIST, CIFAR and House pricing dataset.
  The experimental results demonstrate that the proposed CMP algorithms are more effective than existing attack mechanisms, such as Arjun's attack and label flipping attack.
    More specifically, our original CMP can achieve a high rate of attacker's accuracy  ($\approx90\%$).   For instance, the aggregated model can be manipulated successfully by the CMP under the Krum, and then wrongly identify a given digit $9$ as $8$ in MNIST.
    Meanwhile, our CMP with approximated constraint achieves a rate of $87\%$ in terms of the attacker's accuracy and a $73\%$ complexity reduction relative to the original CMP.
   Furthermore, the proposed blind CMP algorithm can also achieve a rate of $76\%$ in terms of the attacker's accuracy.
\end{itemize}

The rest of this paper is organized as follows. Section~\ref{sec:prelim} introduces the related background.
A detailed description of the proposed algorithms is given in Section~\ref{sec:targeted_att}.
Section~\ref{sec:experi_set} introduces the experimental setup.
In Section~\ref{sec:perfor_eva}, the experimental results are presented and discussed.
Finally, this paper is concluded in Section~\ref{sec:conclu}. A summary notations can be seen in Tab.~\ref{tab:summ_notations}.

\begin{table}[ht]\caption{Summary of Main Notations}
\centering
\begin{tabular}{cl}
\toprule[0.8pt]
\textbf{Notation}&\textbf{Description}\\
\toprule[0.8pt]
$\boldsymbol{\mathcal{U}}$& The set of all clients\\
\hline
$U$& Total number of all clients\\
\hline
$\boldsymbol{\mathcal M}$& The set of compromised clients\\
\hline
$M$& The number of compromised clients\\
\hline
$\boldsymbol{\mathcal{B}}$& The set of benign clients\\
\hline
$B$& The number of benign clients\\
\hline
$\mathcal C_i$& The $i$-th client\\
\hline
$\mathcal D_i$& The dataset held by the client $\mathcal C_i$\\
\hline
$\boldsymbol{\mathcal D}$& The set of all clients' datasets\\
\hline
$|\cdot|$& The cardinality of a set\\
\hline
$t$& The index of the $t$-th communication round\\
\hline
$T$& The number of communication rounds\\
\hline
$\boldsymbol{\theta}$& The vector of model parameters\\
\hline
\multirow{2}*{\shortstack{$\boldsymbol{\theta}^{t}$}}& Global parameters aggregated from local\\
&parameters at the $t$-th communication round\\
\hline
$\boldsymbol{\theta}_{i}^{t}$& Local training parameters of the $i$-th client\\
\hline
${\boldsymbol{\widehat{\theta}}}_{i}^{t}$& Local training parameters after attack's crafting\\
\hline
$\boldsymbol{\theta}^{*}$& The optimal parameters that minimize $F(\boldsymbol{\theta})$\\
\hline
$F(\boldsymbol{\theta})$& Global loss function\\
\hline
$F_{i}(\boldsymbol{\theta})$& Local loss function from the $i$-th client\\
\toprule[0.8pt]
\end{tabular}
\label{tab:summ_notations}
\end{table}

\section{Preliminaies}\label{sec:prelim}
In this section, we will present preliminaries and related background knowledge on FL and model poisoning attack.
\subsection{Federated Learning}
As a kind of decentralized training frameworks~\cite{Mcmahan2016Communication}, FL can preserve clients' private information by its unique distribution learning mechanism.
In details, all participants $\mathcal{C}_{i}$, $\forall i \in \boldsymbol{\mathcal{U}}$, $\boldsymbol{\mathcal{U}}=\{1,2,\ldots,U\}$ only need to share the same learning objective and model structure, where the central server sends the current global model parameters to all clients in each communication round.


Then, each client uploads the model parameters after the local training procedure based on the shared global model and local datasets $\mathcal{D}_{i}$, $\forall i \in \boldsymbol{\mathcal{U}}$, and then all uploaded models will be averaged by the server as the current global model, which is expressed as
\begin{equation}\label{equ:aggregation}
\boldsymbol{\theta}^{t}=\sum\limits_{i\in\boldsymbol{\mathcal{U}}}p_{i}\boldsymbol{\theta}_{i}^{t},
\end{equation}
where $\boldsymbol{\theta}^{t}$ is the global model at the $t$-th communication round, $\boldsymbol{\theta}_{i}^{t}$ is the uploaded model of $i$-th client at the $t$-th communication round, $p_{i}=\vert \mathcal{D}_{i}\vert/\vert \boldsymbol{\mathcal{D}}\vert$ and $\boldsymbol{\mathcal{D}}=\sum_{i\in\boldsymbol{\mathcal{U}}}\cup \mathcal{D}_{i}$.
At the server, the goal is to learn a model over data that resides at the $U$ associated clients.
Formally, this FL task can be expressed as
\begin{equation}\label{equ:global_loss}
\boldsymbol{\theta}^{*}=\mathop{\arg\min}_{\boldsymbol{\theta}\in \boldsymbol{\Theta}}{F(\boldsymbol{\mathcal{D}}, \boldsymbol{\theta})},
\end{equation}
where $\boldsymbol{\Theta}$ represents the domain of legal models, $F(\boldsymbol{\mathcal{D}}, \boldsymbol{\theta})=\sum_{i\in \boldsymbol{\mathcal{U}}}p_{i}F(\mathcal{D}_{i}, \boldsymbol{\theta})$ and $F(\mathcal{D}_{i},\cdot)$ is the local objective function of the $i$-th client.

\subsection{Model Poisoning Attack}\label{sec:Mod_up_pois}

\begin{figure}[ht]
\centering
\includegraphics[width=3.0in,angle=0]{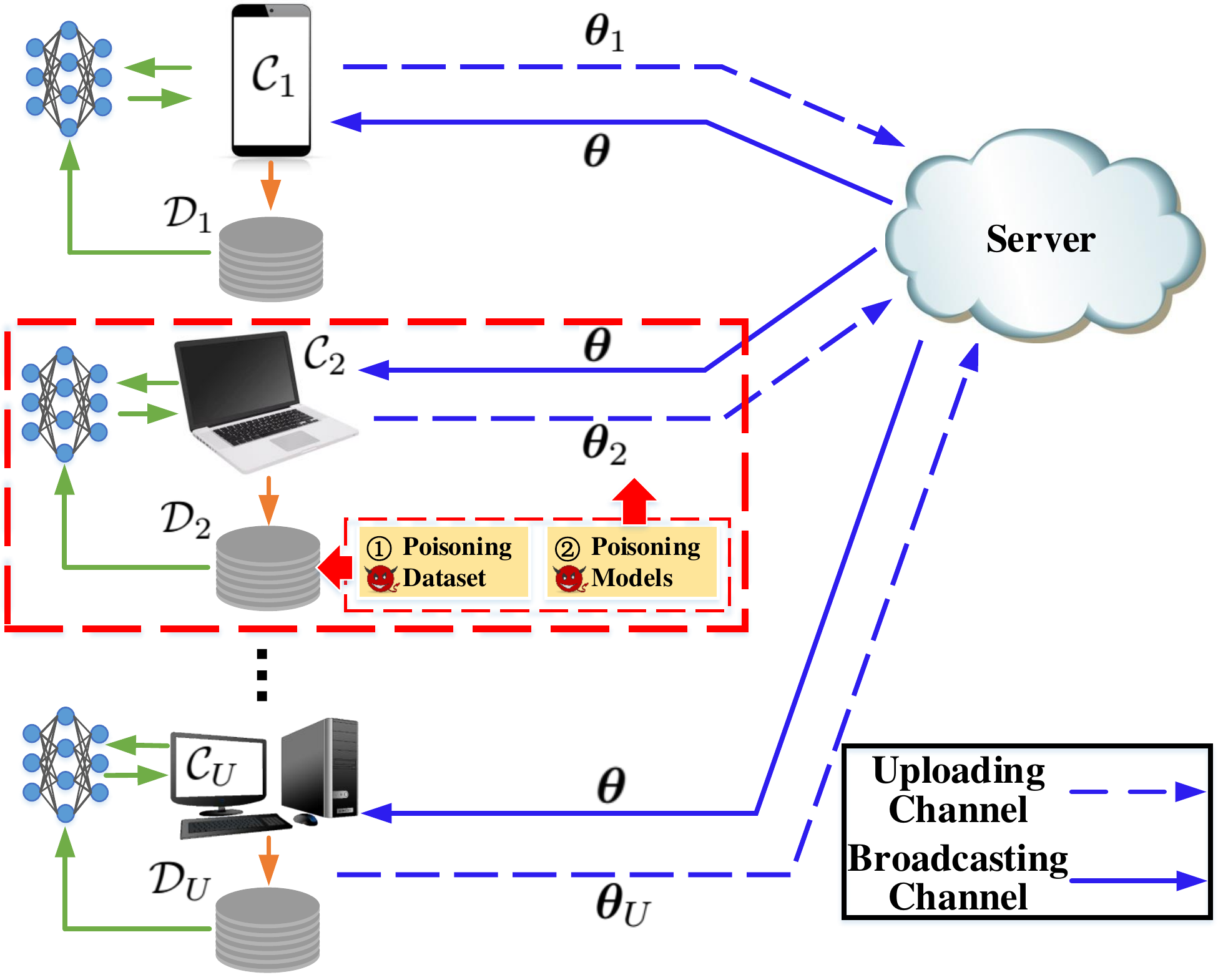}
\caption{Data poisoning attacks vs. model poisoning attacks.}
\label{fig:TrainingModel}
\end{figure}

In FL, model poisoning attack is a natural and powerful attack class~\cite{kairouz2019advances}, where an attacker can directly manipulate updates to the central server.
Fig.~\ref{fig:TrainingModel} shows a high level of model poisoning attack compared with the data poisoning attack.
We will focus on settings where an attacker controls some number of clients and craft the uploaded model parameters.
This can result in convergence to sub-optimal models, or even lead to divergence.
If the attacker has access to the model or non-compromise clients (updates and datasets), they may be able to craft their outputs to have similar variances and magnitudes as the correct model updates, making them difficult to detect.

\textbf{Attacker's Goal:} The goal of an attacker can be classified into two categories: untargeted attacks and targeted attacks.
Of particular importance to untargeted model makes the learnt model unusable and eventually lead to denial-of-service attacks~\cite{fang2019local}.
For instance, an attacker may perform such attacks to its competitor's FL system.
In targeted poisoning attacks, the learnt model produces attacker-desired predictions for particular testing examples, e.g., predicting spams as non-spams and predicting attacker-desired labels for testing examples with a particular trojan trigger (these attacks are also known as backdoor/trojan attacks~\cite{bagdasaryan2018backdoor}).

\textbf{Attacker's Background Knowledge:} In this work, we assume that the attacker may know the global model, local training datasets and local models on the compromised clients. Therefore, we characterize the attacker's background knowledge in three scenarios:
\begin{itemize}
\item \textbf{Full Knowledge Background.} In the case of full knowledge background, the attacker knows the local training datasets and local models of all the clients as well as the aggregation rule. In particular, the attacker could know the aggregation rule in various scenarios. For instance, the service provider may make the aggregation rule public in order to increase transparency and trust of the FL system.
\item \textbf{Partial Knowledge Background.} In the case of partial knowledge background, besides the aggregation rule, the attacker only knows the global model, local training datasets and local models of the compromised clients.
\item \textbf{No Knowledge Background.} In the case of no knowledge background, the attacker does not know the aggregation rule. However, since the attacker controls the compromised clients, it knows their local training datasets and local models.
\end{itemize}

\textbf{Participant Collusion:} An important axis to evaluate in the context of specific federated settings is the capability of participant collusion. In training-time attacks, there may be various attackers compromising various numbers of clients. Intuitively, the attackers may be more effective if they are able to coordinate their poisoned updates than if they each acted individually. Collusion may not happen in `real time' (within-update), but rather across model updates (cross-update collusion).

\section{Proposed Covert Model Poisoning}\label{sec:targeted_att}
In this section, we will propose CMP algorithms with the aim of targeted attacks under various knowledge backgrounds.
\subsection{Problem Formulation for CMP}
In the model poisoning attack, the attacker may directly control a number of compromised clients and manipulate their uploaded models to influence the behavior of the learning algorithm according to predefined goals.
We assume that there exists $M$ clients being compromised by an attacker, and it will directly alter the outputs of these clients to bias the learned model towards to the objective $F_{A}(\cdot)$.
We also define $B$ as the number of benign clients and we have $M+B=U$.
We define $\boldsymbol{\mathcal M}$ as the set of these compromised clients, $\boldsymbol{\mathcal{B}}$ as the set of benign clients, and $\boldsymbol{\mathcal{U}}$ as the set of all clients, where $\boldsymbol{\mathcal M}\subseteq \boldsymbol{\mathcal{U}}$ and $\boldsymbol{\mathcal{B}}= \boldsymbol{\mathcal{U}}/\boldsymbol{\mathcal M}$.
In details, in each communication round, each benign client computes a local parameter vector $\boldsymbol{\theta}_{i}$, $\forall i \in \boldsymbol{\mathcal{B}}$, but each compromised client provides an unreliable parameter vector $\boldsymbol{\widehat{\theta}}_{i'}$, $\forall i' \in \boldsymbol{\mathcal M}$.
With a specific aggregation rule and all uploaded models, the server can update the global model.
A traditional aggregation rule is to average the local model parameters as the global model parameters.
For example, considering the mean aggregation rule, the aggregated model parameter $\boldsymbol{\widehat{\theta}}$ can be expressed as
\begin{equation}\label{equ:malicious_aggregation}
\begin{aligned}
\boldsymbol{\widehat{\theta}}=\sum_{i \in \boldsymbol{\mathcal{B}}}p_{i}\boldsymbol{\theta}_{i}+\sum_{i' \in \boldsymbol{\mathcal M}} p_{i'} \boldsymbol{\widehat{\theta}}_{i'},
\end{aligned}
\end{equation}
Due to the existence of the unreliable parameter vectors from compromised clients, the performance of the aggregated model $\boldsymbol{\widehat{\theta}}^{t}$ may be bad.
However, the goal of the attacker is usually to find a set of $M$ local poisoning models that minimizes the objective function $F_{A}(\cdot)$ when they are uploaded to the server.
Hence, we can formulate the attacker's objective of each communication round as the following optimization problem:
\begin{equation}\label{equ:attack optimization}
\begin{aligned}
\boldsymbol{\widehat{\theta}}^{*}_{\boldsymbol{\mathcal M}} = &\mathop{\arg\min}_{\boldsymbol{\widehat{\theta}}_{i'}\subseteq \boldsymbol{\Theta}, i'\in\boldsymbol{\mathcal M}} F_{A}\left(\boldsymbol{\widehat{\theta}}\right),\\
\mathbf{s.t.}\quad &\boldsymbol{\widehat{\theta}} = \mathcal{A}(\boldsymbol{\widehat{\theta}}_{i'};\boldsymbol{\theta}_{i}), \forall i'\in\boldsymbol{\mathcal M},i\in\boldsymbol{\mathcal{B}},
\end{aligned}
\end{equation}
where $\mathcal{A}$ represents the aggregation rule and $\boldsymbol{\widehat{\theta}}^{*}_{\boldsymbol{\mathcal M}}\triangleq\{\boldsymbol{\widehat{\theta}}_{i'}^{*}|i'\in\boldsymbol{\mathcal M}\}$ represents the optimal poisoning models.
\subsection{CMP for Mean Aggregation}
In the previous work, the attacker's goal is to make the aggregated model minimize the attacker's objective function $F_A(\cdot)$, not the legitimate clients' objective function $F(\cdot)$, e.g., mislead a spam filter to pass certain types of spam emails.
If the server adopts the mean aggregation rule as Eq.~\eqref{equ:aggregation}, substituting the constrain Eq.~\eqref{equ:malicious_aggregation} into the optimization objective, and then we can obtain the following optimization problem:
\begin{equation}\label{equ:attack optimization_mean}
\begin{aligned}
\boldsymbol{\widehat{\theta}}^{*}_{\boldsymbol{\mathcal M}} = &\mathop{\arg\min}_{\boldsymbol{\widehat{\theta}}_{i'}\subseteq \boldsymbol{\Theta}, i'\in\boldsymbol{\mathcal M}} F_{A}\left(\boldsymbol{\widehat{\theta}}\right),\\
&\text{s.t.}\,\, \boldsymbol{\widehat{\theta}}=\sum_{i \in \boldsymbol{\mathcal{B}}}p_{i}\boldsymbol{\theta}_{i}+\sum_{i' \in \boldsymbol{\mathcal M}} p_{i'} \boldsymbol{\widehat{\theta}}_{i'},
\end{aligned}
\end{equation}
where $\boldsymbol{\Theta}$ is the feasible domain of the FL training models.
The attacker's objective function $F_{A}(\boldsymbol{\widehat{\theta}})$ has been typically computed on a specific target model $\boldsymbol{\widehat{\theta}}^{*}$.
In this example, we may define $F_{A}(\boldsymbol{\widehat{\theta}})=\Vert \boldsymbol{\widehat{\theta}}-\boldsymbol{\widehat{\theta}}^{*} \Vert^{2}$ with an appropriate norm.
If the attacker has the full knowledge of this system, we know that the solution of this optimization is available by solving the optimization problem in Eq.~\eqref{equ:attack optimization_mean} directly.

However, if the attacker has the partial knowledge as described in Section~\ref{sec:Mod_up_pois}, the optimal solution can be obtained in the following theorem.
\begin{theorem}\label{theorem:mean_aggre}
With a certain target $F_{A}(\boldsymbol{\widehat{\theta}})$ and $M$ compromised clients under the mean aggregation rule in FL, the crafted model can be calculated by
\begin{equation}
\begin{aligned}
\boldsymbol{\widehat{\theta}}_{i}=\frac{1}{\sum\limits_{i\in\boldsymbol{\mathcal M}}p_{i}}\left(\boldsymbol{\widehat{\theta}}^{*}+\left(\frac{2}{\sum\limits_{i\in\boldsymbol{\mathcal M}}p_{i}}-1\right)\sum_{i\in\boldsymbol{\mathcal M}}p_{i}\boldsymbol{\theta}_{i}\right), \forall i\in \boldsymbol{\mathcal M},
\end{aligned}
\end{equation}
and the loss function value can be expressed as
\begin{equation}
F_{A}\left(\boldsymbol{\widehat{\theta}}\right)= \left(\frac{2}{\sum\limits_{i\in\boldsymbol{\mathcal M}}p_{i}}-1\right)^2\left\Vert\sum_{i\in\boldsymbol{\mathcal M}}p_{i}\boldsymbol{\theta}_{i}\right\Vert^2.
\end{equation}
\end{theorem}
\begin{IEEEproof}
See Appendix~\ref{appendix:mean_aggre}.
\end{IEEEproof}

From~\textbf{Theorem~\ref{theorem:mean_aggre}}, we can obtain a solution for the attacker by estimating the local models of benign clients.
This estimation method will also be used in the following algorithms.
If the sever applies a robust aggregation rule (e.g., Krum~\cite{Blanchard2017} and trimmed mean~\cite{Dong2019Byzantine}), the optimization of the term $\boldsymbol{\widehat{\theta}}^{*}_{\boldsymbol{\mathcal M}}$ in Eq.~\eqref{equ:attack optimization} will be more complicated. In the following subsection, we will propose attacking algorithms against the Krum aggregation.
\subsection{CMP for Krum Aggregation with Full and Partial Knowledge Background}
%
Krum~\cite{Blanchard2017} selects one of the $U$ local models that is the most similar to other models as the global model.
The intuition is that even if the selected local model is from a compromised client, its impact may be constrained since it is similar to other local models possibly from benign clients.
In details, for each local model $\boldsymbol{\theta}_{i}$, the server computes the $U-M-2$ local models that are the closest to $\boldsymbol{\theta}_{i}$ with respect to Euclidean distance.
Moreover, the server computes the sum of the distances between $\boldsymbol{\theta}_{i}$ and its closest $U-M-2$ local models.
Krum selects the local model with the smallest sum of distances as the global model.
When $M < \frac{U-2}{2}$ , Krum has theoretical guarantees for the convergence for certain objective functions.

Without a loss of generality, we assume the first $M$ clients are compromised.
Our directed deviation goal is to craft local models $\boldsymbol{\widehat{\theta}}_{1},\ldots,\boldsymbol{\widehat{\theta}}_{M} $ of the compromised clients.
Recall that Krum selects one local model as the global model in each communication round.
Suppose $\boldsymbol{\widehat{\theta}}$ is the selected local model in the current communication round.
Our goal is to craft the $M$ compromised local models such that the local model selected by Krum has the optimal solution to minimize $F_{A}(\boldsymbol{\widehat{\theta}})$.
Therefore, we can make Krum select a certain crafted local model (e.g., $\boldsymbol{\widehat{\theta}}_{1}$ without a loss of generality) via crafting the $M$ compromised local models.
Then, we aim to solve the optimization problem in Eq.~\eqref{equ:attack optimization for krum} by $\boldsymbol{\widehat{\theta}}=\boldsymbol{\widehat{\theta}}_{1}$.
Therefore, under the Krum rule $\mathcal{A}_\text{krum}$, our optimization problem can be expressed as
\begin{equation}\label{equ:attack optimization for krum}
\begin{aligned}
\boldsymbol{\widehat{\theta}}^{*}_{\boldsymbol{\mathcal M}} = &\mathop{\arg\min}_{\boldsymbol{\widehat{\theta}}_{i'}\subseteq \boldsymbol{\Theta}, i'\in\boldsymbol{\mathcal M} }F_{A}\left(\boldsymbol{\widehat{\theta}}\right),\\
&\text{s.t.}\,\, \boldsymbol{\widehat{\theta}} = \mathcal{A}_\text{krum}(\boldsymbol{\widehat{\theta}}_{i'};\boldsymbol{\theta}_{i}), \forall i'\in\boldsymbol{\mathcal M},i\in\boldsymbol{\mathcal{B}}.
\end{aligned}
\end{equation}
We know that the output of Krum rule is an integer programming problem.
This optimization requires solving a bilevel problem in which the outer optimization amounts to minimize the attacker's objective and the known dataset by the attacker, while the inner optimization corresponds to the aggregation rule on all received models.
Since solving this problem is highly complex, previous work~\cite{jagielski2018manipulating} has exploited gradient-based optimization, along with the idea of implicit differentiation.
Under these conditions, it is possible to apply a gradient descent strategy to obtain a (possibly) local minimum of the optimization problem of Eq.~\eqref{equ:attack optimization for krum} in an iterative manner.

Our proposed CMP algorithm against Krum is given as~\textbf{Algorithm~\ref{alg:local poisoning attack}}.
First, we define the known dataset and local models by the attacker as $\boldsymbol{\mathcal{D}}_{\text{att}}$ and $\boldsymbol{\theta}_{\text{att}}$, respectively.
In the case of full knowledge background, we can obtain that $\boldsymbol{\mathcal{D}}_{\text{att}}=\boldsymbol{\mathcal{D}}_{\boldsymbol{\mathcal{U}}}$ and $\boldsymbol{\theta}_{\text{att}}=\boldsymbol{\theta}_{\boldsymbol{\mathcal{U}}}$, where $\boldsymbol{\theta}_{\boldsymbol{\mathcal U}}\triangleq\{\boldsymbol{\theta}_{i}|i\in\boldsymbol{\mathcal U}\}$ and $\boldsymbol{\mathcal{D}}_{\boldsymbol{\mathcal U}}\triangleq\{\mathcal{D}_{i}|i\in\boldsymbol{\mathcal U}\}$.
Correspondingly, in the case of partial knowledge background, we have $\boldsymbol{\mathcal{D}}_{\text{att}}=\boldsymbol{\mathcal{D}}_{\boldsymbol{\mathcal M}}$, $\boldsymbol{\theta}_{\text{att}}=\boldsymbol{\theta}_{\boldsymbol{\mathcal M}}$, where $\boldsymbol{\theta}_{\boldsymbol{\mathcal M}}\triangleq\{\boldsymbol{\theta}_{i}|i\in\boldsymbol{\mathcal M}\}$ and $\boldsymbol{\mathcal{D}}_{\boldsymbol{\mathcal M}}\triangleq\{\mathcal{D}_{i}|i\in\boldsymbol{\mathcal M}\}$.
With the different degrees of the knowledge background, the algorithm optimizes all compromised models $\boldsymbol{\theta}_{\boldsymbol{\mathcal M}}$ in each communication round, by updating their feature vectors according to a given direction obtained by the gradient descent strategy.
Therefore, at the $t$-th communication round, the update rule can be expressed as:
\begin{equation}
\begin{aligned}
\boldsymbol{\widehat{\theta}}^{t}_{1,k+1} = \Pi_{\boldsymbol{\Theta}}\left(\boldsymbol{\widehat{\theta}}^{t}_{1,k}-\eta_{k}\nabla_{\boldsymbol{\widehat{\theta}}_{1}} F_{A}(\boldsymbol{\widehat{\theta}}_{1,k}^{t})\right),
\end{aligned}
\end{equation}
where $k$ represents the iteration when optimizing Eq.~\eqref{equ:attack optimization for krum} and $\Pi_{\boldsymbol{\Theta}}(\cdot)$ is a projection operator to project $\boldsymbol{\theta}$ onto the feasible domain $\boldsymbol{\Theta}$, to handle bounded feature values.
Note that this update step should also enforce $\boldsymbol{\widehat{\theta}}^{t}_{1,k+1}$ to lie within the feasible domain $\boldsymbol{\Theta}$, which can be typically achieved through the robust aggregation rule.
Then, in order to achieve the goal of participant collusion, this algorithm will obtain the updated $\boldsymbol{\widehat{\theta}}^{t}_{i,k+1}$, $i = 2,3,\ldots,M$ by adding slight noises (Gaussian noises with zero mean and $\sigma$ standard deviation) to $\boldsymbol{\widehat{\theta}}^{t}_{1,k+1}$.
Each noise vector should be clipped by the clipping threshold $\varepsilon$.
After updating the crafted models, this algorithm will check whether Krum selects $\boldsymbol{\widehat{\theta}}^{t}_{1,k+1}$ as the global model. if not, then we decrease the step size $\eta$ with a decay parameter $\lambda$. We repeat this process until Krum selects $\boldsymbol{\widehat{\theta}}^{t}_{1,k+1}$ or $\eta$ is smaller than a certain threshold $\varsigma$.

\begin{algorithm}
\caption{Original Covert Model Poisoning}
\label{alg:local poisoning attack}
\begin{algorithmic}[1]
\Require $\boldsymbol{\mathcal{D}}_{\text{att}}=\boldsymbol{\mathcal{D}}_{\boldsymbol{\mathcal{U}}}$ (full knowledge), $\boldsymbol{\mathcal{D}}_{\text{att}}=\boldsymbol{\mathcal{D}}_{\boldsymbol{\mathcal M}}$ (partial knowledge), $\sigma$, $\eta_{0}$, $\varepsilon$ and $\lambda$.
\State $t\leftarrow 0$ (communication round counter)
\While {$t<T$}
\State The compromised clients craft models as follows:
\State $\boldsymbol{\theta}^{t}_{\text{att}}=\boldsymbol{\theta}^{t}_{\boldsymbol{\mathcal{U}}}$ (if full knowledge) and
\State $\boldsymbol{\theta}^{t}_{\text{att}}=\boldsymbol{\theta}^{t}_{\boldsymbol{\mathcal M}}$ (if partial knowledge)
\State $\boldsymbol{\widehat{\theta}}_{1,k}^{t}\leftarrow \boldsymbol{\theta}^{t}$ and $k\leftarrow 0$ (iteration counter)
\Repeat
\State $\boldsymbol{\widehat{\theta}}^{t}_{1,k+1} \leftarrow \Pi_{\boldsymbol{\Theta}}\left(\boldsymbol{\widehat{\theta}}^{t}_{1,k}-\eta_{k}\nabla_{\boldsymbol{\widehat{\theta}}_{1}} F_{A}(\boldsymbol{\widehat{\theta}}_{1,k}^t)\right)$
\For {$i = 2,3,\ldots,M$}
\State $\boldsymbol{n}_{i}\leftarrow \mathcal{N}(0,\sigma)$
\State $\boldsymbol{\widehat{\theta}}^{t}_{i,k+1}\leftarrow \boldsymbol{\widehat{\theta}}^{t}_{1,k+1}+\varepsilon\boldsymbol{n}_{i}/\Vert \boldsymbol{n}_{i}\Vert$
\EndFor
\State $\boldsymbol{\widehat{\theta}}_{k+1}^{t} \leftarrow \mathcal{A}_\text{krum}(\boldsymbol{\widehat{\theta}}^{t}_{i'};\boldsymbol{\theta}^{t}_{i}), \forall i'\in\boldsymbol{\mathcal M},i\in\boldsymbol{\mathcal{B}}$
\If {$\boldsymbol{\widehat{\theta}}^{t}_{1,k+1}\neq\boldsymbol{\widehat{\theta}}_{k+1}^{t}$}
\State $\eta_{k+1}\leftarrow\lambda\eta_{k}$
\State $\boldsymbol{\widehat{\theta}}^{t}_{1,k+1} \leftarrow \boldsymbol{\widehat{\theta}}^{t}_{1,k}$
\EndIf
\State $k \leftarrow k+1$
\Until{$\eta_{k}<\varsigma$}
\EndWhile
\end{algorithmic}
\end{algorithm}

\subsection{CMP for Krum Aggregation with Low Complexity}
The aforementioned algorithm is essentially a standard gradient-ascent algorithm with the integer constraint.
The key challenge of solving the optimization problem is that the constraint of the optimization problem is highly nonlinear and the search space of the local models $\boldsymbol{\widehat{\theta}}_{1},\ldots,\boldsymbol{\widehat{\theta}}_{M}$ is large.
We know that there are $M-1$ compromised clients assisting $\boldsymbol{\widehat{\theta}}_{1}$.
Because $M<\frac{U-2}{2}$, then we know $M-1<\frac{U}{2}-2<\frac{U}{2}-1<U-M-2$.
Therefore, it is necessary for the crafted model $\boldsymbol{\widehat{\theta}}_{1}$ to be close to $U-2M-1$ benign clients' with respect to Euclidean distance.
Consider the differences among $\boldsymbol{\widehat{\theta}}_{i}$, $\forall i\in \boldsymbol{\mathcal M}$, we set an enough small value $\varepsilon$, where $\Vert \boldsymbol{\widehat{\theta}}_{i} - \boldsymbol{\widehat{\theta}}_{j}\Vert \leq \varepsilon$, $\forall i,j\in \boldsymbol{\mathcal M}$.
The sum of the Euclidean distances of the crafted model $\boldsymbol{\widehat{\theta}}_{1}$ can be expressed as $\mathop{\min}\sum_{j\in \boldsymbol{\mathcal{S}}'}\Vert\boldsymbol{\theta}_{j}-\boldsymbol{\widehat{\theta}}_{1}\Vert+(M-1)\cdot\varepsilon$, where $\boldsymbol{\mathcal{S}}'\subseteq \boldsymbol{\mathcal{B}}$ is the subset of $\boldsymbol{\mathcal{B}}$ and $\vert \boldsymbol{\mathcal{S}}'\vert=U-2M-1$.
Intuitively, we can utilize the sum of the Euclidean distances to simplify the Krum rule.
Thus, we make the following approximation:
\begin{equation}
\begin{aligned}
\mathop{\min}_{i\in \boldsymbol{\mathcal{U}}, \boldsymbol{\mathcal{S}}\subseteq \boldsymbol{\mathcal{U}}/i,\atop\vert \boldsymbol{\mathcal{S}}\vert=U-M-2}\sum_{j\in \boldsymbol{\mathcal{S}}}\Vert\boldsymbol{\theta}_{i}-\boldsymbol{\theta}_{j}\Vert
\lessapprox \mathop{\min}_{i\in \boldsymbol{\mathcal{B}}, \boldsymbol{\mathcal{S}}\subseteq \boldsymbol{\mathcal{B}}/i,\atop\vert \boldsymbol{\mathcal{S}}\vert=U-M-2}\sum_{j\in \boldsymbol{\mathcal{S}}}\Vert\boldsymbol{\theta}_{i}-\boldsymbol{\theta}_{j}\Vert,
\end{aligned}
\end{equation}
where $\boldsymbol{\mathcal{S}}\subseteq \boldsymbol{\mathcal{U}}/i$ is the subset of $\boldsymbol{\mathcal{U}}/i$.
Our approximation represents suboptimal solutions to the optimization problem, which means that the attacks based on this approximation may have suboptimal performance.
After this approximation, we can simplify the aforementioned optimization problem as
\begin{equation}\label{equ:op_krum}
\begin{aligned}
\boldsymbol{\widehat{\theta}}^{*}_{1} = &\mathop{\arg\min}_{\boldsymbol{\theta}_{1} \subseteq \boldsymbol{\Theta} }F_{A}\left(\boldsymbol{\widehat{\theta}}^{*},\boldsymbol{\widehat{\theta}}_{1}\right),\\
\text{s.t.}\,\, &\mathop{\min}\sum_{j\in \boldsymbol{\mathcal{S}}'}\Vert\boldsymbol{\widehat{\theta}}_{1}-\boldsymbol{\theta}_{j}\Vert+(M-1)\cdot\varepsilon-E\leq0,\\
&\boldsymbol{\mathcal{S}}'\subseteq \boldsymbol{\mathcal{B}},\vert \boldsymbol{\mathcal{S}}'\vert=U-2M-1,
\end{aligned}
\end{equation}
where
\begin{equation}
\begin{aligned}
E =\mathop{\min}_{i\in \boldsymbol{\mathcal{B}}, \boldsymbol{\mathcal{S}}\subseteq \boldsymbol{\mathcal{B}}/i,\atop\vert \boldsymbol{\mathcal{S}}\vert=U-M-2}\sum_{j\in \boldsymbol{\mathcal{S}}}\Vert\boldsymbol{\theta}_{i}-\boldsymbol{\theta}_{j}\Vert.\\
\end{aligned}
\end{equation}
In order to simplify the constraints of $\boldsymbol{\mathcal{S}}'$, we can transform Eq.~\eqref{equ:op_krum} into a mixed optimization problem as
\begin{equation}\label{equ:final optimization}
\begin{aligned}
\boldsymbol{\widehat{\theta}}^{*}_{1} = &\mathop{\arg\min}_{\boldsymbol{\widehat{\theta}}_{1} \subseteq \boldsymbol{\Theta} }F_{A}\left(\boldsymbol{\widehat{\theta}}_{1}\right),\\
&\text{s.t.}\,\,\mathop{\min}_{\alpha}\sum_{j \in \boldsymbol{\mathcal{B}}}\alpha_{j}\Vert\boldsymbol{\theta}_{j}-\boldsymbol{\widehat{\theta}}_{1}\Vert\\
&\quad\quad\quad+(M-1)\cdot\varepsilon-E\leq0,\\
&\sum_{j \in \boldsymbol{\mathcal{B}}}\alpha_{j} = U-2M-1, \alpha_{j}\in \{0,1\}.
\end{aligned}
\end{equation}
The bilevel optimization problem in~\eqref{equ:final optimization} is NP hard in general.
Specifically, we require the attack space $\boldsymbol{\Theta}$ to be differentiable (e.g. the attacker can change the local models in $\boldsymbol{\Theta}$ for aggregation).
We know that the objective $F_{A}(\cdot)$ of the attacker is usually convex.
In the following, we will present an efficient solution for a broad class of local model poisoning attacks.

We can note that without the integer constraint, the Lagrangian method offers an effective solution for this optimization problem. Therefore, we decompose this problem into two problems $\textbf{P}_1$ and $\textbf{P}_2$.
The problem $\textbf{P}_1$ can be expressed as
\begin{equation}
\begin{aligned}
\boldsymbol{\widehat{\theta}}^{\star}_{1} = &\mathop{\arg\min}_{\boldsymbol{\widehat{\theta}}_{1} \subseteq \boldsymbol{\Theta} }F_{A}\left(\boldsymbol{\widehat{\theta}}_{1}\right),\\
&\text{s.t.}\,\,\sum_{j \in \boldsymbol{\mathcal{B}}}\alpha_{j}\Vert\boldsymbol{\theta}_{j}-\boldsymbol{\widehat{\theta}}_{1}\Vert+(M-1)\cdot\varepsilon-E\leq0.
\end{aligned}
\end{equation}
The solution of $\textbf{P}_1$ can be expressed by $\boldsymbol{\widehat{\theta}}^{\star}_{1} = h(\alpha)$. Therefore, the optimization problem $\textbf{P}_2$ can be given by:
\begin{equation}
\begin{aligned}
\boldsymbol{\widehat{\theta}}^{*}_{1} = &\mathop{\arg\min}_{\alpha }F_{A}\left(h(\alpha)\right)\\
&\text{s.t.}\,\,\sum_{j \in \boldsymbol{\mathcal{B}}}\alpha_{j} = U-2M-1\\
&\quad\,\, \alpha_{j}(\alpha_{j}-1) = 0, \forall j\in \boldsymbol{\mathcal{B}}.
\end{aligned}
\end{equation}
Specifically, the Lagrangian function of the problem $\textbf{P}_1$ can be written as
\begin{equation}
\begin{aligned}
\mathcal{L}(\mathbf{\alpha},\lambda)&=F_{A}\left(\boldsymbol{\widehat{\theta}}_{1}\right)+\lambda\cdot{\Bigg[}\sum_{j \in \boldsymbol{\mathcal{B}}}\alpha_{j}\Vert\boldsymbol{\theta}_{j}-\boldsymbol{\widehat{\theta}}_{1}\Vert\\
&\quad+(M-1)\cdot\varepsilon-E{\Bigg{]}},
\end{aligned}
\end{equation}
Based on the Karush-Kuhn-Tucker (KKT) conditions, the model $\boldsymbol{\widehat{\theta}}^{\star}_{1}$ and the optimal Lagrangian multiplier $\lambda$ should satisfy the following equation set:
\begin{equation}\label{equ:equ_set}
\left\{
\begin{aligned}
&\frac{\partial F_{A}\left(\boldsymbol{\widehat{\theta}}_{1}\right)}{\partial \boldsymbol{\widehat{\theta}}_{1}}+\lambda\cdot\sum\limits_{j \in \boldsymbol{\mathcal{B}}}\frac{\alpha_{j}(\boldsymbol{\widehat{\theta}}^{\star}_{1}-\boldsymbol{\theta}_{j})}{\Vert \boldsymbol{\widehat{\theta}}_{1}^{\star}-\boldsymbol{\theta}_{j}\Vert}=0, \lambda>0\\
&\sum_{j \in \boldsymbol{\mathcal{B}}}\alpha_{j}\Vert\boldsymbol{\widehat{\theta}}_{1}^{\star}-\boldsymbol{\theta}_{j}\Vert+(M-1)\cdot\varepsilon-E = 0.\\
\end{aligned}
\right.
\end{equation}

Based on the equation set~\eqref{equ:equ_set}, we can obtain the model $\boldsymbol{\widehat{\theta}}^{\star}_{1}$ for the problem $\textbf{P}_1$.
Conventional nonlinear optimization often has high local searching ability but low global searching ability.
The genetic algorithm uses survival of the fittest as a method to achieve a good solution for its optimization problem.
Combining the genetic algorithm, we can escape from local minima and obtain satisfied results.
By this approximation, we can obtain low complexity CMP algorithm based on~\textbf{Algorithm~\ref{alg:local poisoning attack}}.

Furthermore, we can find that how to choose an initial model is crucial for the proposed algorithm
Therefore, we derive an initial model for the optimization problem in \textbf{Algorithm~\ref{alg:local poisoning attack}}.
Formally, we have the following theorem.
\begin{theorem}\label{theorem:simple_result}
With the attacker's objective function $F_{A}(\boldsymbol{\widehat{\theta}}^{*},\boldsymbol{\widehat{\theta}}_{1})=\Vert \boldsymbol{\widehat{\theta}}^{*}-\boldsymbol{\widehat{\theta}}_{1}\Vert^{2}$, $\boldsymbol{\widehat{\theta}}_{\emph{init}}$ is properly an initial model for CMP algorithm in each communication round as follows:
\begin{equation}
\begin{aligned}
&\boldsymbol{\widehat{\theta}}_{\emph{init}} = \Xi+\left(\boldsymbol{\widehat{\theta}}^{*}-\Xi\right)\\
&\left(\left\Vert\Xi\right\Vert^2-\frac{1}{B}\sum\limits_{j \in \boldsymbol{\mathcal{B}}}\Vert\boldsymbol{\theta}_{j}\Vert^2+\Lambda\right)^{\frac{1}{2}}{\LARGE/}\left\Vert\boldsymbol{\widehat{\theta}}^{*}-\Xi\right\Vert,
\end{aligned}
\end{equation}
where $\Lambda = -\frac{U-2M-2}{2}-(M-1)\varepsilon+E$, $\Xi = \frac{1}{B}\sum\limits_{j \in \boldsymbol{\mathcal{B}}}\boldsymbol{\theta}_{j}$, and $\boldsymbol{\widehat{\theta}}^{*}$ is the adversarial objective model.
\end{theorem}
\begin{IEEEproof}
See Appendix~\ref{appendix:simple_result}.
\end{IEEEproof}

From~\textbf{Theorem \ref{theorem:simple_result}}, we can note that if $\Lambda$ is larger, the result will be closer to $\boldsymbol{\widehat{\theta}}^{*}$.
The intuition is that a larger $\Lambda$ means a loose constraint and leads to a better solution for this optimization problem.

\subsection{CMP with No Knowledge Background}
In this subsection, we consider the CMP with no knowledge background (CMP-NKB) based on the aforementioned algorithm.
When the attacker does not know the aggregation rule, this attacker can only adjust the crafted models via the feedback of the aggregation rule.


\textbf{Algorithm~\ref{alg:covert_targeted_poisoning}} outlines our proposed CMP-NKB. At the $t$-th communication round, the server broadcasts the aggregated global model $\boldsymbol{\theta}^{t}$ to all clients.
The benign clients respectively train the parameters by using local databases with preset termination conditions.
After completing the local training, the $i$-th client, $\forall i$, will upload the local parameters $\boldsymbol{\theta}^{t}_i$ to the server for aggregation.
However, different from these benign clients, the attacker aims to craft effective models to evade the robust aggregation rule instead of local training.
In the CMP-NKB, when the global model has been received at the $t$-th communication round, the attacker may calculate the distance between the $\boldsymbol{\theta}^{t}$ and $\widehat{\boldsymbol{\theta}}_{1}^{t}$.
Using this distance and threshold $\xi$, we can obtain the result whether our crafted models evade the robust aggregation rule successfully.
If the result is positive, the attacker will enhance the poisoning degree by increasing the step size of the label flipping training.
Meanwhile, the attacker need to weaken the poisoning degree by decreasing the step size.

The poisoning problem under the robust aggregation can be viewed as a global game between two players, i.e., a learner and an attacker.
The game captures the interactions on a network of a FL training model including $U-M$ benign clients and a centralized server, and an attacker with $M$ compromised clients.
However, if we treat each compromised client as an independent attacker, then the global game can be treated as a multi-agent problem, which can be valuable future work and solved by the multi-agent reinforcement learning algorithms.

\begin{algorithm}
\caption{Original Covert Model Poisoning with No Knowledge Background (CMP-NKB)}
\label{alg:covert_targeted_poisoning}
\begin{algorithmic}[1]
\Require $\boldsymbol{\mathcal{D}}_{\text{att}}=\boldsymbol{\mathcal{D}}_{\boldsymbol{\mathcal M}}$, $\eta^{0}$ and $\lambda$.
\State $t\leftarrow 0$ (communication round counter)
\While {$t<T$}
\State The server broadcasts the aggregated model
$\boldsymbol{\theta}^{t}$
\State The compromised clients craft models as follows:
\If {$\Vert \boldsymbol{\widehat{\theta}}^{t}_{1}-\boldsymbol{\theta}^{t}\Vert \leq \xi$}
\State $\eta^{t+1}\leftarrow\lambda\eta^{t}$
\Else
\State $\eta^{t+1}\leftarrow\eta^{t}/\lambda$
\EndIf
\State $\boldsymbol{\widehat{\theta}}^{t}_{1} \leftarrow \Pi_{\boldsymbol{\Theta}}\left(\boldsymbol{\theta}^{t}-\eta^{t}\nabla_{\boldsymbol{\theta}^{t}} F_{A}(\boldsymbol{\theta}^{t};\boldsymbol{\mathcal{D}}_{\text{att}})\right)$
\For {$i = 2,3,\ldots,M$}
\State $\boldsymbol{n}_{i}\leftarrow \mathcal{N}(0,\sigma)$
\State $\boldsymbol{\widehat{\theta}}^{t}_{i}\leftarrow \boldsymbol{\widehat{\theta}}^{t}_{i}+\varepsilon\boldsymbol{n}_{i}/\Vert \boldsymbol{n}_{i}\Vert$
\EndFor
\State All clients upload the local models to the server
\State The server aggregate uploaded models by
\State a certain aggregation rule
\State $t \leftarrow t+1$
\EndWhile
\end{algorithmic}
\end{algorithm}
\section{Experimental Setup}\label{sec:experi_set}
In this section, we implemented our attacks using Pytorch.
We trained all of the models on a server equipped with three Tesla P100 PCIe and each with $16$ GB of memory.
We evaluate the effectiveness of our proposed attack methods using multiple datasets and learning models in different scenarios, e.g., the impact of different parameters and known vs. different aggregation rules.
\subsection{Datasets}
Our experiments use three real datasets:

1) \textbf{House pricing dataset.} House pricing dataset is used to predict house sale prices as a function of predictor variables such as square
footage, number of rooms, and location~\cite{Kaggle2017House}. In total, it includes 1460 houses and 81 features. We preprocess by onehot encoding all categorical features and normalize numerical features, resulting in 275 total features;

2) \textbf{MNIST.} MNIST is a dataset of handwritten digits consists of $60000$ training examples and $10000$ testing examples~\cite{Lecun1998Gradient} formatted as 28$\times$28 size gray scale images;

3) \textbf{CIFAR-$10$.}  CIFAR-$10$ consists of $60000$ color images in $10$ object classes such as deer, airplane, and dog with $6000$ images included per class. The complete dataset is pre-divided into $50000$ training images and $10000$ test images.

We normalize each numerical dimension to $[0,1]$. In addition, we also map labels to numbers such that the distance between two points with different labels is no smaller than the distance between points with the same label. In training and testing SVM, we map one label to `$1$' and the rest to `$-1$'. Each data point has a unit weight.

\subsection{Machine Learning Models}
Our experiments evaluate four supervised learning models including linear regression, support vector machine (SVM), multi-layer perceptron (MLP) and conventional neural network (CNN).
1) Linear regression (LR), which is performed with SGD on the House pricing dataset. We use the normalized cost as the following loss function. The considered aggregation rules have theoretical guarantees for the error rate of LR classifier. we conduct the house pricing prediction task by LR;
2) SVM, which is trained on the IPUMS-US dataset. In this model, the hinge loss function is applied. We conduct experiments using the SVM classifier to predict whether the digit is even or odd;
3) MLP, which is conducted on the standard MNIST dataset. For MNIST, We use a simple feedforward deep neural network with ReLU units and softmax of 10 classes (corresponding to the 10 digits) with the cross-entropy loss;
4) CNN, which has $2$ convolutional layers with dropout and is applied for the CIFAR-$10$ dataset. For CIFAR-$10$, we also use softmax of 10 classes with the cross-entropy loss.

Our machine learning architecture does not necessarily achieve the smallest error rates for the considered datasets, as our goal is not to search for the best learning architecture.
Our goal is to show that our attack methods can increase the testing error rates of the machine learning classifiers or bias the learned model towards the attack's objective.
\subsection{Benchmarks}
Furthermore, we compare existing attacks with our proposed methods, which are detailedly described as follows.
1) Gaussian attack. Specifically, for each compromised client, we sample a noise vector from the Gaussian distribution and add it on the parameter of the local model on the compromised client. We use this Gaussian attack to show that crafting compromised local models randomly can not effectively attack the Byzantine-robust aggregation rules;
2) Label flipping attack. This is a data poisoning attack that does not require knowledge of the training data distribution. On each compromised worker device, this attack flips the label of each training instance;
3) Fang's Full knowledge attack or partial knowledge attack~\cite{fang2019local}. These attacks manipulate the local model parameters on compromised worker devices during the learning process against the Byzantine robust FL.
4) Arjun' attack~\cite{bhagoji2018analyzing}. This attack considered the non-colluding malicious clients and designed an alternating minimization strategy, which alternately optimizes for the training loss and the adversarial objective.
\subsection{Performance Metrics}
For the House Pricing prediction, we evaluate the performance by the normalized cost (test loss).
We use this dataset for the experiment of untargeted attack and a larger value of test loss means a better attack performance.
For the MNIST, if we consider the case of untargeted attack, we will use the error rate of the FL model to evaluate our CMP. Specifically, if our CMP can achieve a high error rate, it can attack the FL system effectively.
We also apply the MNIST and CIFAR-$10$ in the case of targeted attack.
In this scenario, we will use attacker's accuracy (predicting the attacker-desired labels by testing data) to evaluate our CMP.
Moreover, if the Krum is adopted in the FL system, we denote Successful Attacking Rate as the rate that compromised models are selected by the Krum in each communication.
\subsection{Parameter Setting}
We describe parameter settings for the FL and our attack methods. We record each experiment for $50$ trials and report the average results.
In our FL system, the number of total clients is set to $50$, the number of compromised client is set from $0$ to $10$ and the degree of non-independent-and-identically-distributed (non-i.i.d.) is set in the range of $[0,1]$.
The targeted attack flips the labels of the MNIST or CIFAR-$10$ dataset with attacker-desired labels, which are shown in Tab.~\ref{Tab:Targeted_labels}.

\begin{table*}[htp]
\caption{Description of the targeted attack (original labels and attacker-desired labels).}
\centering
\begin{tabular}{|m{2.1cm}<{\centering}|c|c|c|c|c|c|c|c|c|c|}
\hline
\multicolumn{11}{|c|}{MNIST/CIFAR-$10$}\\
\hline
Original Labels&$0$/Airplane&$1$/Automobile&$2$/Bird&$3$/Cat&$4$/Deer&$5$/Dog&$6$/Frog&$7$/Horse&$8$/Ship&$9$/Truck\\
\hline
Attacker-desired Labels&$9$/Truck&$0$/Airplane&$1$/Automobile&$2$/Bird&$3$/Cat&$4$/Deer&$5$/Dog&$6$/Frog&$7$/Horse&$8$/Ship\\
\hline
\end{tabular}\label{Tab:Targeted_labels}
\end{table*}

Inspired by the above optimization for the targeted attack, we can craft local models of compromised clients achieve untargeted attacks via solving the similar optimization, which only differs form the objective function.
We can use the original objective function of the benign client and maximize it as the objective.
As same as the targeted attack, we can utilize the same assumptions and approximations, and then solve it by the gradient ascent strategy.
\section{Performance Evaluation}\label{sec:perfor_eva}
In this section, we start by presenting our results for the stand-alone scenario, followed by our results for the FL scenario.
We perform the original CMP under full and partial knowledge backgrounds, denoted by CMP-FKB-Orgcontr and CMP-PKB-Orgcontr in our experiments, respectively.
Correspondingly, the CMP attack with low complexity under full and partial knowledge backgrounds are named as CMP-FKB-Simcontr and CMP-PKB-Simcontr, respectively.

In Section~\ref{subsec:house_pricing} and Section~\ref{subsec:par_classify}, we consider the untargeted attack, where the attacker aim to destroy the FL model.
In Section~\ref{sbusec:handwr_reco} and Section~\ref{subsec:visual_resul}, we attack the MLP and CNN with MNIST or CIFAR-$10$ datasets with the proposed targeted attacking algorithms, respectively, where original labels and attacker-desired labels are shown in Tab.~\ref{Tab:Targeted_labels}.
\subsection{House Pricing Prediction}\label{subsec:house_pricing}
\begin{figure}[ht]
\centering
\includegraphics[width=2.8in,angle=0]{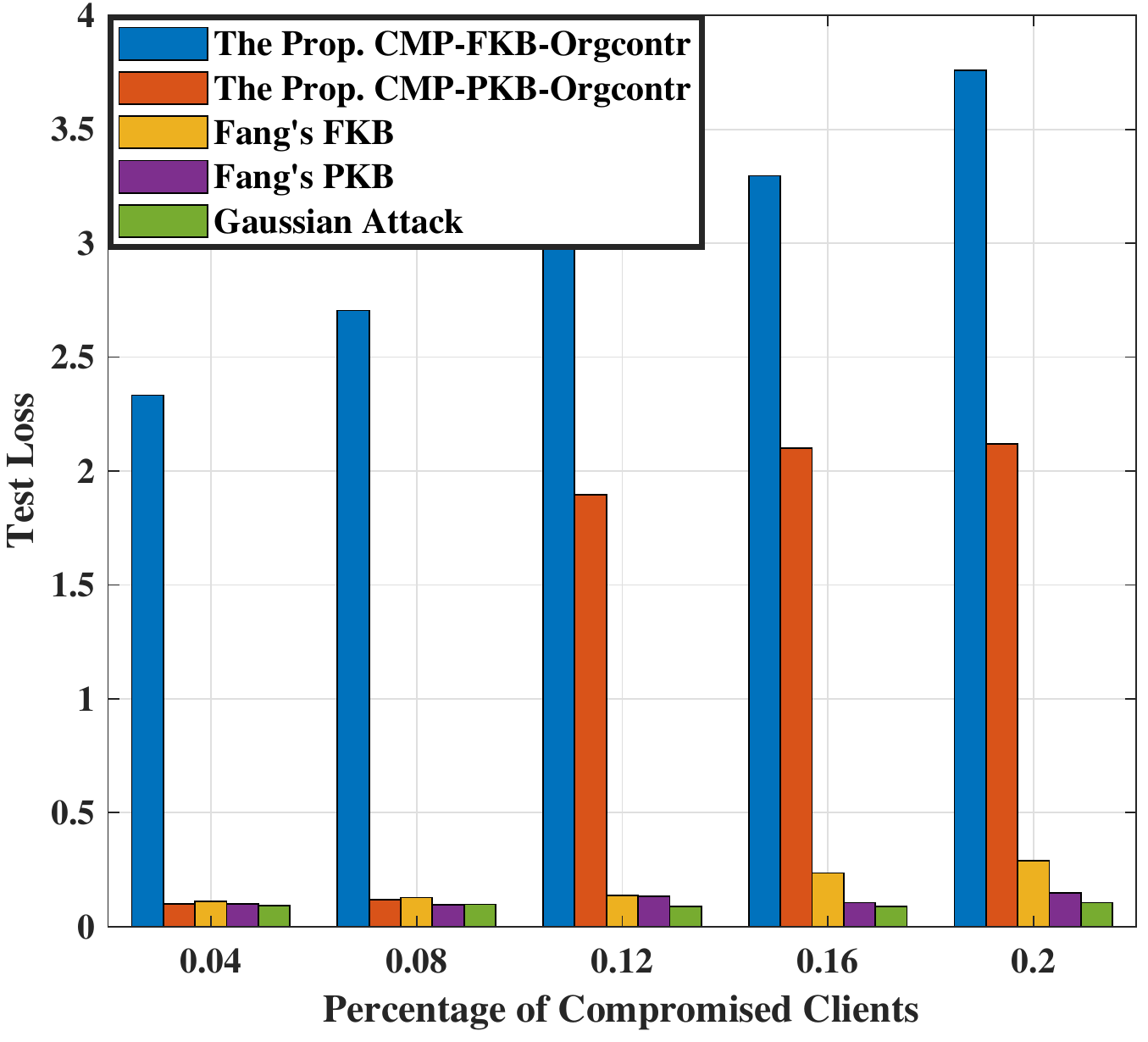}
\caption{The loss function of our proposed untargeted attacking algorithms i.e. the prop. CMP-PKB and the prop. CMP-FKB, against Krum aggregation compared with existing works under various percentages of compromised clients with $p=0.5$ and $T=30$.}
\label{fig:loss_fedreg}
\end{figure}

\begin{figure}[ht]
\centering
\includegraphics[width=2.8in,angle=0]{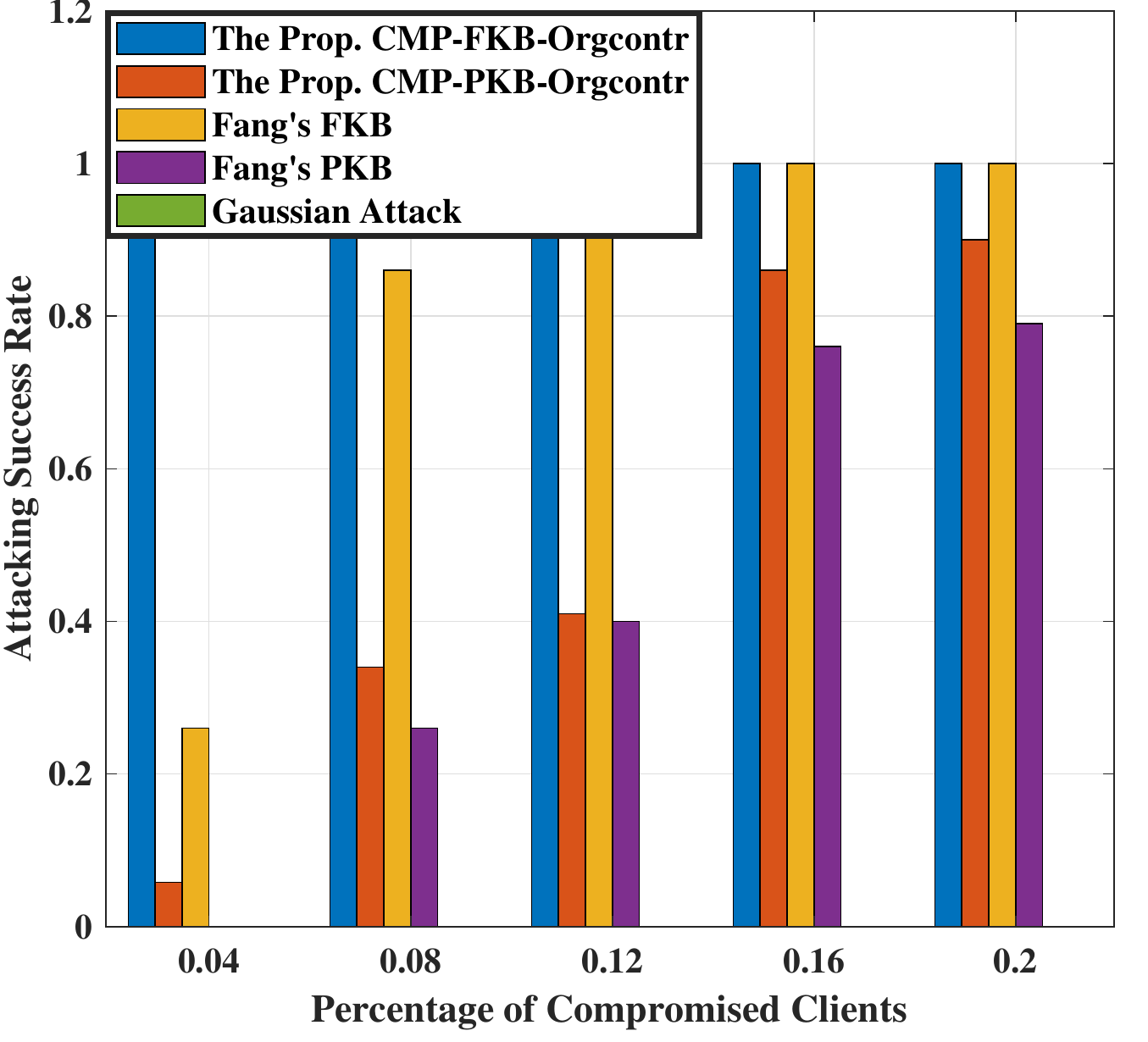}
\caption{The attack success rate of our proposed untargeted attacking algorithms i.e. the prop. CMP-PKB and the prop. CMP-FKB, against Krum aggregation compared with existing works under various percentages of compromised clients with $p=0.5$ and $T=30$.}
\label{fig:asr_fedreg}
\end{figure}
In the first scenario, we evaluate the performance of the house pricing prediction task by LR with the normalized cost.
Figs.~\ref{fig:loss_fedreg} and ~\ref{fig:asr_fedreg} show the loss function value and attack success rate using our untargeted attack methods under various percentages of compromised clients compared with existing works, respectively.

The results in Fig.~\ref{fig:loss_fedreg} show that our attacks are effective and substantially outperform existing attacks.
If the attacker has the full knowledge, our proposed attack can damage this LR model completely while the robust aggregation rule is existing.
However, our proposed attack algorithm will have less effect compared with the full knowledge attack.
We can also find that when the percentage of compromised clients is larger, our proposed attack will have a deeper effect, which is in line with the intuition.
In Fig.~\ref{fig:asr_fedreg}, we show the attack success rate of our proposed attacks against the Krum rule compared with existing works.
Our proposed attacks outperform other methods in both full knowledge and partial knowledge attacks.

Furthermore, Our proposed attack algorithms increase the error rates significantly as we compromise more clients and Gaussian attacks have no notable impact on the error rates.
In Fig.~\ref{fig:asr_fedreg}, we can note that the attack success rate under the partial knowledge background increases with the percentage of compromised clients.

\subsection{Parity Classifier using MNIST}\label{subsec:par_classify}
In this subsection, we conduct experiments using the SVM classifier to predict whether the digit is even or odd.
In Tab.~\ref{Tab:Untargeted_Svm}, we show the error rate and the attack success rate of our proposed untargeted attacking methods against Krum aggregation compared with existing works, respectively.

First, these results show that our attacks are effective and substantially outperform existing attacks, i.e., our attacks result in higher error rates.
For instance, when the degree of non-i.i.d. is set to $0$, our CMP-FKB attack increases the error rate from $0.287$ to 0.425 (around $48.08\%$ relative increase) compared with Fang's FKB.
Furthermore, our CMP-PKB attack also increases the error rate as well as the attack success rate compared with Fang's PKB under different degrees of non-i.i.d..
Finally, we can note that when the degree of non-i.i.d. is small, the attacker under the partial knowledge background has a large value of the attack success rate.
The intuition is that if the degree of non-i.i.d. is small, the divergence of different local models will be small and the estimation of the unknown models by the compromised models will be accurate.

\begin{table*}[htp]
\caption{The comparison of test accuracy between our proposed untargeted attack algorithms, i.e. the prop. CMP-PKB, the prop. CMP-FKB and existing algorithms on SVM model and MNIST dataset under various degrees of non-i.i.d. with $U=50$, $M=10$ and $T=30$.}
\centering
\begin{tabular}{c||c|c|c|c}
\hline
The Degree of Non-i.i.d.&Approach&Error Rate (\%)&Successful Attacking Rate (\%)& Time Cost (second)\\
\hline\hline
\multirow{5}*{\shortstack{$p=0$}}&The Prop. CMP-FKB-Orgcontr&$42.498$&$100$&$3.357$\\
&The Prop. CMP-PKB-Orgcontr&$22.201$&$92.0$&$1.268$\\
&Fang's FKB&$28.714$&$100$&$0.077$\\
&Fang's PKB&$16.863$&$82.6$&$0.115$\\
&Gaussian Attack&$15.510$&$0.00$&$0.00073$\\
\hline\hline
\multirow{5}*{\shortstack{$p=0.5$}}&The Prop. CMP-FKB-Orgcontr&$43.254$&$100$&$3.338$\\
&The Prop. CMP-PKB-Orgcontr&$22.058$&$98.0$&$1.561$\\
&Fang's FKB&$28.985$&$100$&$0.0737$\\
&Fang's PKB&$18.121$&$76.8$&$0.0531$\\
&Gaussian Attack&$15.783$&$0.00$&$0.00072$\\
\hline\hline
\multirow{5}*{\shortstack{$p=1.0$}}&The Prop. CMP-FKB-Orgcontr&$39.161$&$100$&$3.469$\\
&The Prop. CMP-PKB-Orgcontr&$33.306$&$79.8$&$3.784$\\
&Fang's FKB&$31.510$&$100$&$0.067$\\
&Fang's PKB&$23.306$&$58.6$&$0.010$\\
&Gaussian Attack&$18.930$&$0.00$&$0.00073$\\
\hline
\end{tabular}\label{Tab:Untargeted_Svm}
\end{table*}

\begin{table}[htp]
\caption{The attacker's accuracy of our proposed CMP-FKB against mean aggregation compared with existing works under various degrees of non-i.i.d. with $U=50$, $M=10$ and $T=30$.}
\centering
\begin{tabular}{c||c|c}
\hline
The Degree of  &\multirow{2}*{\shortstack{Approach}}&Attacker's\\
Non-i.i.d.&&Accuracy (\%)\\
\hline\hline
\multirow{3}*{\shortstack{$p=0$}}&The Prop. CMP-FKB&$89.38$\\
&Arjun's Attack&$12.50$\\
&Label Flipping&$4.16$\\
\hline\hline
\multirow{3}*{\shortstack{$p=0.5$}}&The Prop. CMP-FKB&$86.91$\\
&Arjun's Attack&$11.71$\\
&Label Flipping&$3.77$\\
\hline\hline
\multirow{3}*{\shortstack{$p=1.0$}}&The Prop. CMP-FKB&$90.29$\\
&Arjun's Attack&$8.78$\\
&Label Flipping&$3.84$\\
\hline
\end{tabular}\label{Tab:mean_Targeted_NKB}
\end{table}

\begin{table*}[htp]
\caption{The comparison of attacker's accuracy between the proposed targeted attack algorithms, i.e. the Prop. CMP-FKB-Simcontr, the Prop. CMP-PKB-Orgcontr, the Prop. CMP-PKB-Simcontr, the Prop. CMP-PKB-Simcontr, and existing algorithms on MLP model and MNIST dataset, under various numbers of communication rounds with $p = 0.5$, $U=50$ and $M=10$.}
\centering
\begin{tabular}{c||c|c|c|c}
\hline
The Number of &\multirow{2}*{\shortstack{Approach}}&\multirow{2}*{\shortstack{Attacker's Accuracy (\%)}}&\multirow{2}*{\shortstack{Successful Attacking Rate (\%)}}&\multirow{2}*{\shortstack{Time Cost (Second)}}\\
Communication Rounds&&&& \\
\hline\hline
\multirow{6}*{\shortstack{$T=30$}}&The Prop. CMP-FKB-Orgcontr&$87.8$&$33.3$&$63.90$\\
&The Prop. CMP-FKB-Simcontr&$71.9$&$5.6$&$17.19$\\
&The Prop. CMP-PKB-Orgcontr&$36.6$&$4.0$&$14.01$\\
&The Prop. CMP-PKB-Simcontr&$10.4$&$10.0$&$6.76$\\
&Arjun's Attack&$9.6$&$0.00$&$0.015$\\
&Label Flipping&$9.2$&$0.00$&$0.00$\\
\hline\hline
\multirow{6}*{\shortstack{$T=50$}}&The Prop. CMP-FKB-Orgcontr&$90.4$&$33.3$&$63.90$\\
&The Prop. CMP-FKB-Simcontr&$87.0$&$10.0$&$17.19$\\
&The Prop. CMP-PKB-Orgcontr&$54.6$&$4.6$&$14.01$\\
&The Prop. CMP-PKB-Simcontr&$5.0$&$10.0$&$6.76$\\
&Arjun's Attack&$5.2$&$0.00$&$0.015$\\
&Label Flipping&$4.7$&$0.00$&$0.00$\\
\hline\hline
\multirow{6}*{\shortstack{$T=70$}}&The Prop. CMP-FKB-Orgcontr&$86.7$&$100$&$63.90$\\
&The Prop. CMP-FKB-Simcontr&$87.3$&$33.3$&$17.19$\\
&The Prop. CMP-PKB-Orgcontr&$70.0$&$6.4$&$14.01$\\
&The Prop. CMP-PKB-Simcontr&$3.2$&$100$&$6.76$\\
&Arjun's Attack&$3.3$&$0.00$&$0.015$\\
&Label Flipping&$6.3$&$0.00$&$0.00$\\
\hline
\end{tabular}\label{Tab:Targeted_deep}
\end{table*}

\subsection{Handwriting Digits Recognition}\label{sbusec:handwr_reco}
In this subsection, we conduct our experiments using MNIST by the MLP classifier to classify the handwriting digits.
We show the attack accuracy and the attack success rate of our proposed targeted attacking methods against mean, Krum and Trimmed mean aggregation rules compared with existing works, respectively.

In Tab.~\ref{Tab:mean_Targeted_NKB}, we show the attacker's accuracy of our proposed CMP-FKB against mean aggregation compared with existing works under various degrees of non-i.i.d. with $U=50$, $M=10$ and $T=30$.
We can note that the proposed CMP algorithms are more effective than existing attacks and achieve a high attacker' accuracy (above $\% 85$).

In Tab.~\ref{Tab:Targeted_deep}, we show the comparison of attacker's accuracy between our proposed targeted attack algorithms, i.e. The Prop. CMP-FKB-Simcontr,  The Prop. CMP-PKB-Orgcontr, The Prop. CMP-PKB-Simcontr and The Prop. CMP-PKB-Simcontr, and existing algorithms on MLP model and MNIST dataset, under various numbers of communication rounds with $p = 0.5$, $U=50$ and $M=10$.
These results show that the proposed CMP algorithms are effective and substantially outperform existing attacks, such as Arjun's attack and label flipping attack.
Considering Krum, with the original constraint, our proposed algorithms can usually achieve a high attacker's accuracy, especially for a large $T$. For instance, when $T = 50$, our full knowledge attack with the original constraint increases the attack accuracy from $0.052$ to $0.904$ as well as our partial knowledge attack increases it to $0.546$. Meanwhile, using the approximate constraint, our proposed algorithm will have a small time cost but a low attacker's accuracy.

In Tab.~\ref{Tab:krum_Targeted_NKB}, we show the attacker's accuracy of our proposed CMP-NKB against Krum aggregation compared with existing works under various percentages of compromised clients with $p = 0.5$ and $T=300$. We can note that the proposed algorithm also can achieve a high attacker's accuracy, although this attacker only know the information of compromised clients (the datasets of compromised clients and global model parameters).

In order to show the effectiveness of proposed attacks, we also use the Trimmed mean aggregation~\cite{jagielski2018manipulating} in the FL system.
In Tab.~\ref{Tab:Trimmed_Targeted_deep}, we show the attacker's accuracy of CMP-NKB against Trimmed mean aggregation compared with existing works on MLP model and MNIST dataset with $p = 0.5$ and $T=30$.
We can note that CMP-NKB can also achieve a high attacker's accuracy compared with existing attacks, such as Arjun's attack and label flipping attack.

\begin{table}[htp]
\caption{The attacker's accuracy of the proposed CMP-NKB against Krum aggregation compared with existing works on MLP model and MNIST dataset under various percentages of compromised clients with $p = 0.5$, $U=50$ and $T=300$.}
\centering
\begin{tabular}{c||c|c}
\hline
Percentage of  &\multirow{2}*{\shortstack{Approach}}&Attacker's\\
Compromised Clients&&Accuracy (\%)\\
\hline\hline
\multirow{3}*{\shortstack{0.20}}&The Prop. CMP-NKB&$75.72$\\
&Arjun's Attack&$3.58$\\
&Label Flipping&$2.02$\\
\hline\hline
\multirow{3}*{\shortstack{0.16}}&The Prop. CMP-NKB&$40.43$\\
&Arjun's Attack&$1.30$\\
&Label Flipping&$2.15$\\
\hline\hline
\multirow{3}*{\shortstack{0.12}}&The Prop. CMP-NKB&$11.59$\\
&Arjun's Attack&$2.6$\\
&Label Flipping&$5.14$\\
\hline\hline
\multirow{3}*{\shortstack{0.08}}&The Prop. CMP-NKB&$12.63$\\
&Arjun's Attack&$2.93$\\
&Label Flipping&$2.08$\\
\hline\hline
\multirow{3}*{\shortstack{0.04}}&The Prop. CMP-NKB&$9.24$\\
&Arjun's Attack&$2.80$\\
&Label Flipping&$2.21$\\
\hline
\end{tabular}\label{Tab:krum_Targeted_NKB}
\end{table}

\begin{table}[htp]
\caption{The attacker's accuracy of the proposed CMP-NKB against Trimmed mean aggregation compared with existing works on MLP model and MNIST dataset under various degrees of non-i.i.d. with $U=50$, $M=10$ and $T=300$.}
\centering
\begin{tabular}{c||c|c}
\hline
The Degree of &\multirow{2}*{\shortstack{Approach}}&Attacker's\\
Non-i.i.d. &&Accuracy (\%)\\
\hline\hline
\multirow{3}*{\shortstack{$p=0$}}&The Prop. CMP-NKB&$34.56$\\
&Arjun's Attack&$3.67$\\
&Label Flipping&$4.72$\\
\hline\hline
\multirow{3}*{\shortstack{$p=0.5$}}&The Prop. CMP-NKB&$64.82$\\
&Arjun's Attack&$8.71$\\
&Label Flipping&$5.40$\\
\hline\hline
\multirow{3}*{\shortstack{$p=1.0$}}&The Prop. CMP-NKB&$70.23$\\
&Arjun's Attack&$10.52$\\
&Label Flipping&$5.89$\\
\hline
\end{tabular}\label{Tab:Trimmed_Targeted_deep}
\end{table}

\begin{figure*}[ht]
\centering
\includegraphics[width=6.4in,angle=0]{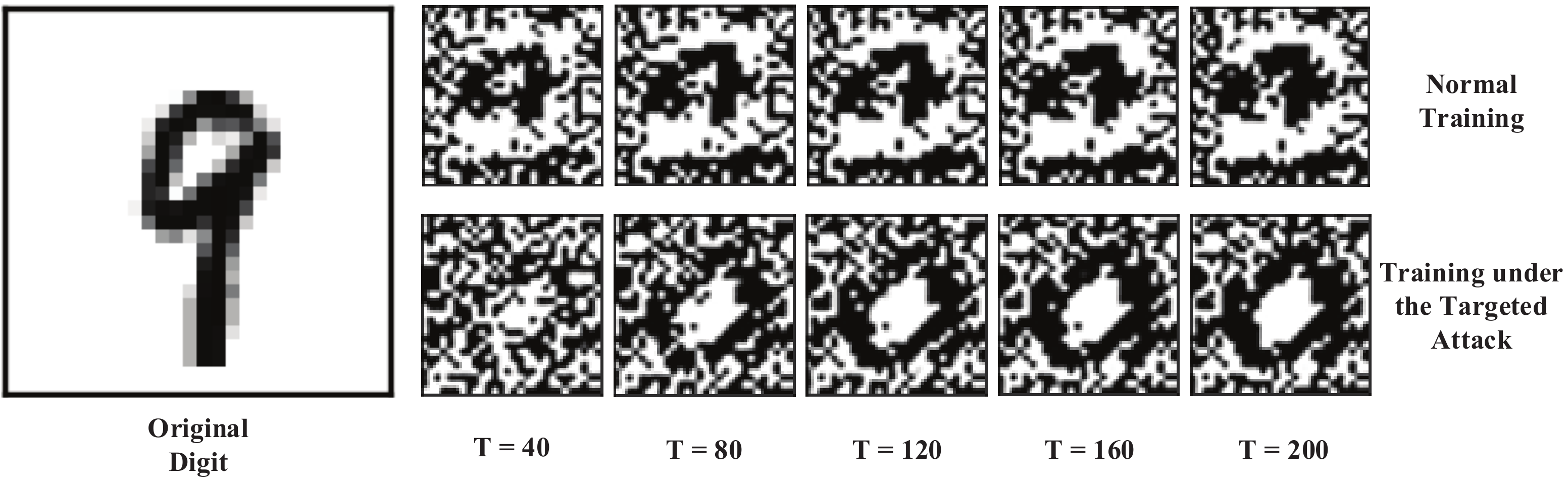}
\caption{The visual results of MLP based FL with the interpretability technique corresponding to the digit $9$ under our proposed CMP-NKB against Krum aggregation at different communication rounds. Basically speaking, the subfigures show a ``typical'' 9 understood by the machine model with normal training and targeted attack.}
\label{fig:mnist_visual_9}
\end{figure*}
\begin{figure*}[ht]
\centering
\includegraphics[width=6.4in,angle=0]{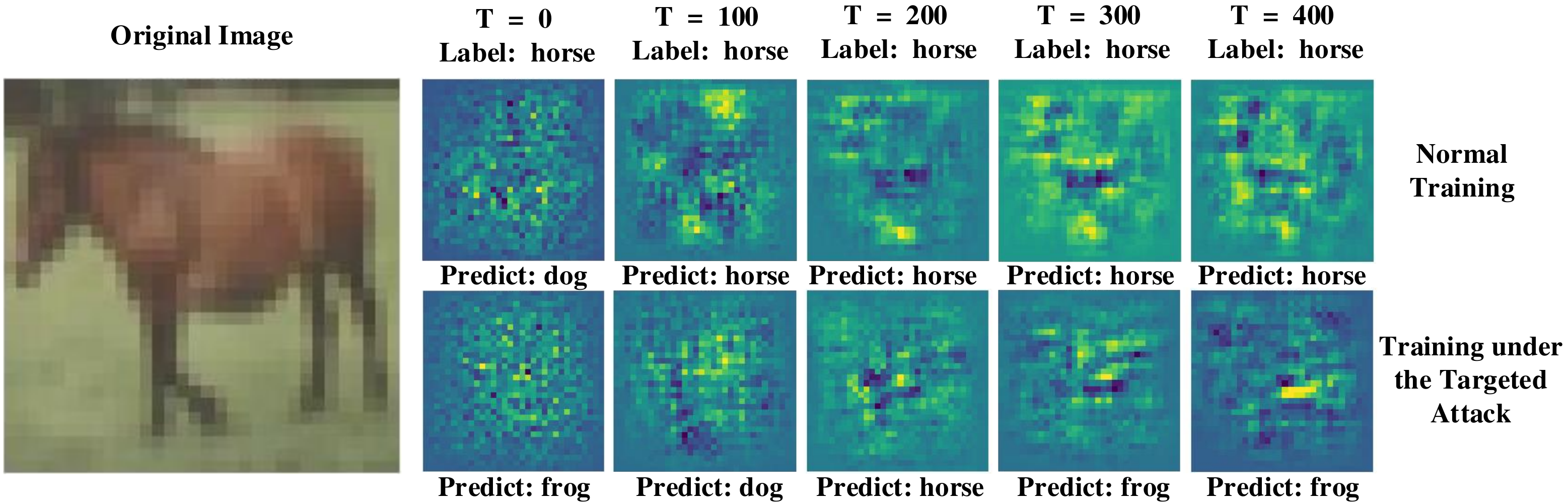}
\caption{The visual results of CNN based FL with the interpretability technique corresponding to the image `horse' under our CMP-NKB against Krum aggregation at different communication rounds. Basically speaking, the subfigures show a ``typical'' horse image understood by the machine model with normal training and targeted attack.}
\label{fig:cifar_visual}
\end{figure*}
\subsection{Visual Results}\label{subsec:visual_resul}
In this subsection, we apply several interpretability techniques, and then provide insights into the internal feature representations of the neural network under the proposed CMP-NKB.
In detail, we use a suite of these techniques to try and discriminate between the behavior of a benign global model and one that has been trained to satisfy the adversarial objective of misclassifying a single example.
Fig.~\ref{fig:mnist_visual_9} compares the outputs of MLP based FL with the interpretability technique corresponding to the digit $9$ for malicious models at different communication rounds.
We also show the results by CNN based FL with CIFAR-10 dataset in Fig.~\ref{fig:cifar_visual}.

In Fig.~\ref{fig:mnist_visual_9}, we can note that the visual results of MLP based FL model corresponding to the digit $9$ using our CMP-NKB against Krum aggregation are similar with the digit $0$.
Especially, with the increasing numbers of communication rounds, the visual results trend to be closer to the digit $0$ instead of $9$, which means that our CMP-NKB can achieve an outstanding attacking performance.
Besides, we also adopt the CNN based FL model with CIFAR-$10$ dataset and show the visual results in~\ref{fig:cifar_visual}.
We can also note that the proposed algorithm can successfully obtain the attacker's goal and make the FL model beneficial to the attacker, i.e. mislead the model to classify the horse as the frog.
\section{Conclusions}\label{sec:conclu}
In this paper, we have performed the systematic study of model poisoning attacks for FL models against defensive aggregation rules, i.e., Krum and Trimmed mean.
We have formulated the model poisoning as an optimization problem by minimizing the Euclidean distance between the manipulated model and designated one, constrained by a defensive aggregation rule.
Then, we have developed CMP algorithms against different defensive aggregation rules according to the solutions of their corresponding optimization problems.
We have also proposed a low complexity CMP algorithm for Krum with a slight performance degradation.
In the case that the attacker does not know the defence mechanism, we have designed a blind CMP algorithm, in which the manipulated model will be adjusted properly according to the aggregated model.
Finally, We have conducted extensive experiments on real-word datasets, i.e., MNIST, CIFAR and House pricing dataset.
The experimental results have demonstrated that the proposed CMP algorithms are more effective than existing attack mechanisms, such as Arjun's attack and label flipping attack.
\bibliographystyle{IEEEtran}
\bibliography{reference}

\clearpage
\appendices
\section{Proof of Theorem~\ref{theorem:mean_aggre}}\label{appendix:mean_aggre}
In the mean aggregation, we know that
\begin{equation}\label{equ:mean_aggre0}
\boldsymbol{\widehat{\theta}}=\sum_{i \in \boldsymbol{\mathcal{B}}}p_{i}\boldsymbol{\theta}_{i}+\sum_{i' \in \boldsymbol{\mathcal M}} p_{i'} \boldsymbol{\widehat{\theta}}_{i'}.
\end{equation}
Substituting~\eqref{equ:mean_aggre0} into $F_{A}(\boldsymbol{\widehat{\theta}})$, we have
\begin{equation}
F_{A}(\boldsymbol{\widehat{\theta}}) = \left\Vert\boldsymbol{\widehat{\theta}}^{*}-\sum_{i \in \boldsymbol{\mathcal{B}}}p_{i}\boldsymbol{\theta}_{i}+\sum_{i' \in \boldsymbol{\mathcal M}} p_{i'} \boldsymbol{\widehat{\theta}}_{i'}\right\Vert^{2}.
\end{equation}
In order to minimize $F_{A}(\boldsymbol{\widehat{\theta}})$, we can obtain
\begin{equation}
\begin{aligned}
\sum_{i' \in \boldsymbol{\mathcal M}} p_{i'}\boldsymbol{\widehat{\theta}}_{i'}= \boldsymbol{\widehat{\theta}}^{*}-\sum_{i\in\boldsymbol{\mathcal{B}}}p_{i}\boldsymbol{\theta}_{i}.
\end{aligned}
\end{equation}
However, the local models of benign clients are unknown for the attacker under the partial knowledge background.
Therefore, The compromised models can be utilized to estimate the unknown models and we have
\begin{equation}\label{equ:approx_models}
\begin{aligned}
\sum_{i\in\boldsymbol{\mathcal M}}p_{i}\boldsymbol{\widehat{\theta}}_{i}&= \boldsymbol{\widehat{\theta}}^{*}-\frac{\sum_{i\in\boldsymbol{\mathcal{U}}}p_{i}-\sum_{i\in\boldsymbol{\mathcal M}}p_{i}}{\sum_{i\in\boldsymbol{\mathcal M}}p_{i}}\sum_{i\in\boldsymbol{\mathcal M}}p_{i}\boldsymbol{\theta}_{i}\\
&= \boldsymbol{\widehat{\theta}}^{*}+\left(\frac{2}{\sum_{i\in\boldsymbol{\mathcal M}}p_{i}}-1\right)\sum_{i\in\boldsymbol{\mathcal M}}p_{i}\boldsymbol{\theta}_{i}.
\end{aligned}
\end{equation}
Note that, we can set
\begin{equation}
\begin{aligned}
\boldsymbol{\widehat{\theta}}_{i}=\frac{1}{\sum_{i\in\boldsymbol{\mathcal M}}p_{i}}\left(\boldsymbol{\widehat{\theta}}^{*}+\left(\frac{2}{\sum_{i\in\boldsymbol{\mathcal M}}p_{i}}-1\right)\sum_{i\in\boldsymbol{\mathcal M}}p_{i}\boldsymbol{\theta}_{i}\right),
\end{aligned}
\end{equation}
which is a solution of~\eqref{equ:approx_models}. Hence, we have
\begin{equation}
\begin{aligned}
F_{A}\left(\boldsymbol{\widehat{\theta}}\right)= \left(\frac{2}{\sum_{i\in\boldsymbol{\mathcal M}}p_{i}}-1\right)^2\left\Vert\sum_{i\in\boldsymbol{\mathcal M}}p_{i}\boldsymbol{\theta}_{i}\right\Vert^2.
\end{aligned}
\end{equation}
This completes the proof. $\hfill\square$
\section{Proof of Theorem~\ref{theorem:simple_result}}\label{appendix:simple_result}
Considering~\eqref{equ:final optimization}, we can further have
\begin{equation}
\begin{aligned}
\mathop{\min}_{\alpha}\sum_{j \in \boldsymbol{\mathcal{B}}}\alpha_{j}\Vert\boldsymbol{\theta}_{j}-\boldsymbol{\widehat{\theta}}_{1}\Vert\leq\mathop{\min}_{\alpha}\sum_{j \in \boldsymbol{\mathcal{B}}}\frac{\alpha_{j}^{2}+\Vert\boldsymbol{\theta}_{j}-\boldsymbol{\widehat{\theta}}_{1}\Vert^2}{2},
\end{aligned}
\end{equation}
and
\begin{equation}
\begin{aligned}
\mathop{\min}_{\alpha}&\sum_{j \in \boldsymbol{\mathcal{B}}}\frac{\alpha_{j}^{2}+\Vert\boldsymbol{\theta}_{j}-\boldsymbol{\widehat{\theta}}_{1}\Vert^2}{2}\\
&=\frac{U-2M-2}{2}+\sum_{j \in \boldsymbol{\mathcal{B}}}\frac{\Vert\boldsymbol{\theta}_{j}-\boldsymbol{\widehat{\theta}}_{1}\Vert^2}{2}.
\end{aligned}
\end{equation}
Then, we can adjust the optimization~\eqref{equ:final optimization} with a stronger constraint as
\begin{equation}
\begin{aligned}
\boldsymbol{\widehat{\theta}}^{*}_{1} = \quad&\mathop{\arg\min}_{\boldsymbol{\widehat{\theta}}_{1} \subseteq \boldsymbol{\Theta} }F_{A}(\boldsymbol{\widehat{\theta}}_{1}),\\
\mathbf{s.t.}\quad &\sum_{j \in \boldsymbol{\mathcal{B}}}\frac{\Vert\boldsymbol{\theta}_{j}-\boldsymbol{\widehat{\theta}}_{1}\Vert^2}{2}+\frac{U-2M-2}{2}+(M-1)\cdot\varepsilon\\
&\quad-E\leq0.
\end{aligned}
\end{equation}
With an auxiliary model $\boldsymbol{\widehat{\theta}}^{*}$, we can obtain an initial solution of~\eqref{equ:final optimization} using $F_{A}(\boldsymbol{\widehat{\theta}}^{*},\boldsymbol{\widehat{\theta}}_{1})=\Vert \boldsymbol{\widehat{\theta}}_{1}-\boldsymbol{\widehat{\theta}}^{*}\Vert^{2}$.
Fortunately, the Lagrangian method offers an effective solution for~\eqref{equ:final optimization}.
Based on the KKT conditions, we can obtain the following solution for $\boldsymbol{\widehat{\theta}}^{*}_{1}$:
\begin{equation}
\begin{aligned}
\left\Vert\boldsymbol{\widehat{\theta}}_{1}-\frac{1}{B}\sum_{j \in \boldsymbol{\mathcal{B}}}\boldsymbol{\theta}_{j}\right\Vert^2-\frac{1}{B^2}\left\Vert\sum_{j \in \boldsymbol{\mathcal{B}}}\boldsymbol{\theta}_{j}\right\Vert^2+\frac{1}{B}\sum_{j \in \boldsymbol{\mathcal{B}}}\Vert\boldsymbol{\theta}_{j}\Vert^2\\
+\frac{2}{B}\left(\frac{U-2M-2}{2}+(M-1)\cdot\varepsilon-E\right)=0.
\end{aligned}
\end{equation}
Therefore, we can obtain the initial model as
\begin{equation}
\begin{aligned}
&\boldsymbol{\widehat{\theta}}_{\text{init}} = \frac{1}{B}\Xi+\left(\boldsymbol{\widehat{\theta}}^{*}-\frac{1}{B}\Xi\right)\\
&\left(\frac{1}{B^2}\left\Vert\Xi\right\Vert^2-\frac{1}{B}\sum\limits_{j \in \boldsymbol{\mathcal{B}}}\Vert\boldsymbol{\theta}_{j}\Vert^2-\Lambda\right)^{\frac{1}{2}}{\LARGE/}\left\Vert\boldsymbol{\widehat{\theta}}^{*}-\frac{1}{B}\Xi\right\Vert,
\end{aligned}
\end{equation}
where $\Lambda = \frac{U-2M-2+2(M-1)\varepsilon-2E}{B}$ and $\Xi = \sum\limits_{j \in \boldsymbol{\mathcal{B}}}\boldsymbol{\theta}_{j}$.
This completes the proof. $\hfill\square$
\end{document}